\documentclass{article}
\usepackage{ijcai26}

\usepackage{times}
\usepackage[dvipsnames,table]{xcolor}
\usepackage{latexsym}

\usepackage[T1]{fontenc}
\usepackage[utf8]{inputenc}

\usepackage{amsmath}
\usepackage{amssymb}
\usepackage{amsthm}

\usepackage{graphicx}
\usepackage{booktabs}
\usepackage{array}
\usepackage{multirow}
\usepackage{adjustbox}
\usepackage{tabularx}

\usepackage{algorithm}
\usepackage{algpseudocode}

\usepackage{enumitem}
\setlist{nosep}

\usepackage{microtype}
\usepackage{subcaption}

\usepackage{listings}
\usepackage[switch]{lineno}
\usepackage{comment}
\usepackage{url}
\usepackage[font=small]{caption}

\title{Addressing Overcommitment in the Reasoning of Gendered Economic Memes under Multimodal Ambiguity}

\author{
    Author Name
    \affiliations
    Affiliation
    \emails
    email@example.com
}

\author{
Kushal Kanwar$^1$
\and
Dushyant Singh Chauhan$^2$\and
Kapil Rana$^{3}$\and \\
Gopendra Vikram Singh$^{4}$\and
Nils Lukas$^2$\\
\affiliations
$^1$Jaypee University of Information Technology\\
$^2$Mohamed bin Zayed University of Artificial Intelligence\\
$^3$Thapar Institute of Engineering \& Technology\\
$^4$Dr. B. R. Ambedkar National Institute of Technology Jalandhar\\
\emails
kushalneo@gmail.com,
dushyant.chauhan@mbzuai.ac.ae,
kapilrana.iitrpr@gmail.com,
gopendra.99@gmail.com,
nils.lukas@mbzuai.ac.ae
}

\begin{document}

\maketitle

\begin{abstract}
Multimodal meme understanding is increasingly used to analyze socially sensitive content, yet existing models often exhibit biased behavior when interpreting economic dependence and social roles under ambiguity. Many memes express economic relationships through sparse text or symbolic visual cues, providing insufficient evidence for gendered attribution. In such underspecified settings, models tend to rely on pretraining correlations, leading to hallucinated and stereotypical economic role assignments. In this work, we study gendered economic dependence in image-text memes through the lens of contextual sufficiency and identify \emph{epistemic overcommitment}-inferring roles without adequate evidence-as a primary source of bias. We propose \textbf{CGER-Net}, a context-grounded multimodal framework that estimates whether the input provides sufficient evidence for gendered economic reasoning and applies evidence-gated inference to enable confident attribution when cues are explicit while favoring principled abstention otherwise. We evaluate CGER-Net on \textbf{EconMeme-GE}, a curated dataset of image-text memes annotated as \textit{Men}, \textit{Women}, \textit{Neutral}, or \textit{Ambiguous}. Across strong contemporary multimodal baselines, CGER-Net reduces Gender Overcommitment Rate by up to \textbf{44\%} on ambiguous instances while maintaining comparable accuracy on unambiguous cases. Human evaluation further shows that \textbf{79\%} of generated rationales are judged as epistemically aligned with the available evidence. These results highlight the importance of modeling when \emph{not} to infer for reliable and responsible multimodal analysis.
\end{abstract}

\section{Introduction}

Multimodal Memes that express sentiments about labor, capital, and financial agency are common.  These memes are being increasingly examined by large multimodal models. Although these models perform well on common benchmarks, they tend to malfunction in socially sensitive settings.  This is particularly the case when information regarding economic agents is vague and implicit. During pretraining, models learn spurious correlations and draw gendered inferences that are socially plausible but not based in reality.

\paragraph{Ambiguity and Hallucinated Economic Inference.}
Unlike the inaccuracies that are obvious for factual hallucinations, the attributions of a hallucinated economic role are subtler and can go unnoticed as they fall in line with socially prevalent stereotypes.  As a result, when models are made to choose one out of multiple gendered categories, they overcommit even when the epistemically appropriate answer is I don’t know. Existing benchmarks and modeling approaches do not commonly differentiate true neutrality from insufficient context to infer, thereby rewarding confident but unwarranted inference in-the-wild under ambiguity. 

\paragraph{Why Existing Approaches Fall Short.}
Most bias mitigation methods debias representations or weaken correlations but ignore whether inputs provide sufficient evidence for gendered economic inference, leading to confident predictions in underspecified contexts. We recast this failure as a problem of contextual sufficiency, where models must separate grounded cases from those requiring epistemic restraint and treat ambiguity as a first-class outcome.

We introduce \textbf{CGER-Net}, a context-grounded multimodal framework for image–text memes that predicts gendered economic dependence and generates evidence-consistent rationales. \textbf{CGER-Net} explicitly estimates evidence sufficiency and gates inference accordingly. Rather than suppressing gender reasoning, it determines when such reasoning is epistemically warranted.

\begin{figure*}[!]
	\centering
	\begin{subfigure}[b]{0.22\textwidth}
		\centering
		\includegraphics[width=\linewidth,height=3cm,keepaspectratio]{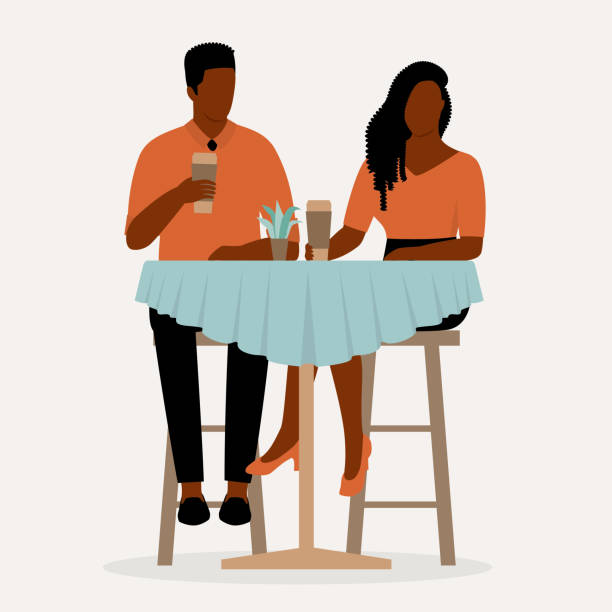}
		\caption{\textbf{Meme Description}: Couple at a restaurant
table \\ \textbf{Label}: Ambiguous\\ \textbf{Text Content}: So who’s paying today? \\ \textbf{Annotation Rationale}: The meme implies financial responsibility but does not specify income source or who is economically responsible.}
		\label{fig:image1}
	\end{subfigure}
	\hfill
	\begin{subfigure}[b]{0.22\textwidth}
		\centering
		\includegraphics[width=\linewidth,height=3cm,keepaspectratio]{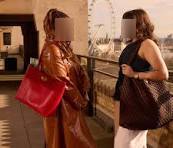}
		\caption{\textbf{Meme Description}: Woman holding shopping bags \\ \textbf{Label}: Women \\ \textbf{Text Content}: My salary, my rules \\ \textbf{Annotation Rationale}: The text explicitly attributes financial agency to the speaker, who is visually identified as a woman.}
		\label{fig:image2}
	\end{subfigure}
	\hfill
	\begin{subfigure}[b]{0.22\textwidth}
		\centering
		\includegraphics[width=\linewidth,height=3cm,keepaspectratio]{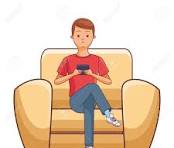}
		\caption{\textbf{Meme Description}: Cartoon of man on couch \\ \textbf{Label}: Men \\ \textbf{Text Content}: Unemployed boyfriend \\ \textbf{Annotation Rationale}: Economic dependence is explicitly stated through unemployment and the subject is clearly male.}
		\label{fig:image3}
	\end{subfigure}
	\hfill
	\begin{subfigure}[b]{0.22\textwidth}
		\centering
		\includegraphics[width=\linewidth,height=3cm,keepaspectratio]{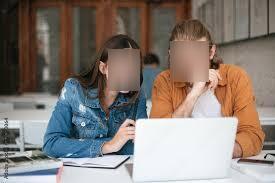}
		\caption{\textbf{Meme Description}: Man and woman working on laptops \\ \textbf{Label}: Neutral \\ \textbf{Text Content}: Teamwork makes it
work \\ \textbf{Annotation Rationale}: Both individuals are depicted as
economically active, with no indication of dependence or gendered financial hierarchy.}
		\label{fig:image4}
	\end{subfigure}
	
	\caption{Illustrative examples from \textbf{EconMeme-GE}, showing image description, textual content, annotation labels, and corresponding human annotation rationales.}
	\label{fig_memes_sample}
\end{figure*}

\paragraph{Contributions.}
Our main contributions are summarized as follows:
\begin{itemize}[nolistsep]
    \item We identify contextual insufficiency as a primary source of hallucinated gendered economic inference in multimodal memes.
    \item We propose a new framework of CGER-Net that models when gendered economic attribution is warranted and when it is not.
    \item We present the first introduction, annotation, and evaluation on EconMeme-GE, a multimodal meme dataset for gendered economic.
    \item Our comprehensive analysis, which includes quantitative, ablation, qualitative, and human-centered approaches, demonstrates that CGER-Net effectively alleviates inappropriate gender assignments in uncertain situations. Furthermore, it achieves this without compromising the essential reasoning used when economic roles are clearly defined.
\end{itemize}

\paragraph{Alignment with UN Sustainable Development Goals (UNSDGs):} This work aligns with \textbf{SDG 10} (Reduced Inequalities) and \textbf{SDG 16} (Peace, Justice and Strong Institutions) by mitigating biased behavior of large language models under uncertainty. By preventing social inferences in ambiguous contexts, it supports fairer automated decision-making and promotes responsible public and institutional adoption of AI.

\section{Related Work}

Gendered economic-dependence memes often provide insufficient evidence to justify attributing dependence to a specific gender. Prior gender-related meme benchmarks largely cast the problem as decidable multimodal classification—e.g., SemEval MAMI \cite{fersini2022semeval}, Hateful Memes \cite{kiela2020hateful}, and MMHS150K \cite{gomez2020exploring} WBMS\cite{kanwar2025beneath} and therefore do not isolate cases where the epistemically appropriate response is restraint (“insufficient evidence”). Similar to these are Building on this line of socially sensitive meme understanding and bias auditing/mitigation for vision-language backbones \cite{radford2021learning,mandal2023multimodal,janghorbani2023multimodal,ravfogel2020null,alabdulmohsin2024clip}, we instead treat \emph{when not to infer} gendered economic roles as the first-class problem by curating EconMeme-GE with an explicit Ambiguous label and introducing a context-sufficiency gate that reduces overcommitment while preserving grounded predictions and rationale alignment when evidence is clear.

\section{Dataset}
\label{sec:dataset}

EconMeme-GE is an economic gendered meme dataset that is annotated for economic gender dependence and epistemic ambiguity. The examples in Figure~\ref{fig_memes_sample} represent the dataset structure and annotation rationale; rationales are human-written, used for evaluation only, not for training the model.

\subsection{Dataset Construction}
We introduce a dataset for image-text memes on economic dependence and social roles and meanings (for dataset availability see, Appendix~\textbf{Dataset Details}). It is common for memes and other forms of discourse to feature explicit and implicit references to money, work, and dependence on other family members, notably a husband or male partner. In order to facilitate the utilization of economically neutral queries that focus on memes (and do not have reference to gender) so as to avoid filtering out memes by gender and thus heavy skew/bias during collection, we try to ensure memes are about similar categories of people (queries include income, unemployed, working, household contribution). Each database sample has meme image and text.

\subsection{Annotation Guidelines}
\label{anno_guidelines}

Three expert annotators (details in Appendix~\textbf{Dataset Details}) categorize memes into one of four mutually exclusive options: \textbf{Men}, \textbf{Women}, \textbf{Neutral}, or \textbf{Ambiguous}. Detailed annotation guidelines are supplied to ensure that judgments are not influenced by personal bias or stereotype-driven reasoning. Annotators should label content as \textbf{Men} or \textbf{Women} only in cases where the meme demonstrates a clear association between the subject and financial agency, either through explicit textual indicators or through compelling visual evidence aligned with the text.

Memes are classified as \textbf{Neutral} when they refer to economic or financial aspects without suggesting the dependence or agency of a gendered subject. Importantly, neutrality does not mean the absence of economic content but the absence of a male or female economic attribution.
Although a meme with a man and a woman may evoke the idea of economic dependence or financial responsibility of one person on the other, the annotator is instructed to choose the label \textbf{Ambiguous} if the meme does not provide enough evidence to ascribe the role to a particular gender. The ambiguous label does not imply a lack of gender in the meme. Rather, the evidence is not strong enough to assign any economic role to either gender.
When annotating, avoid assigning labels based solely on cultural stereotypes or visual clues. This theme runs throughout the annotation; don’t impose a female ascription under an epistemically under-specified context.

\subsection{Annotation Challenges}
\label{annoation_challenge}
There are many non-trivial challenges in annotating economic dependence: \textbf{Implicit economic cues:} Economic roles and dependencies are important and implicit meanings. Although not directly stated, one can deduce their meaning through the sarcasm, irony or cultural generalization present in memes. \textbf{Symbolic visual representations:} Consumer and leisure images, or those that take place within homes such as kitchens or dining rooms can indicate economic contribution (to the household or community) or dependency (on a household). However, they seldom indicate financial agency (in expenditures) or responsibility (for debts). \textbf{Sparse or absent textual information:} Many memes become more epistemically ambiguous because they contain very short or no text at all. This creates a strong temptation for inferring economic roles based on visual appearance and gender. \textbf{Risk of stereotype-driven interpretation:} Characters’ clothing, names or appearance often lead people to assume their genders. Annotators might apply gender-informed socially biased stereotypes when there is no strict directive.

    
    
    

Annotators tend to prefer the better supported interpretation rather than the most plausible one. The annotator selects Ambiguous instead of Neutral when there is no direct evidence.

\subsection{Dataset Characteristics}

The dataset has intentional ambiguities to underline that economic reasoning present in memes is quite underspecified, with background knowledge and experience of individuals playing a role in understanding. Moreover, due to the lack of demographic information, models cannot utilize stereotypes, which is why our dataset does not have external demographic metadata. 

\section{Methodology}
\label{sec:methodology}

\begin{figure*}[!h]
\begin{center}
\includegraphics[width=0.9\textwidth]{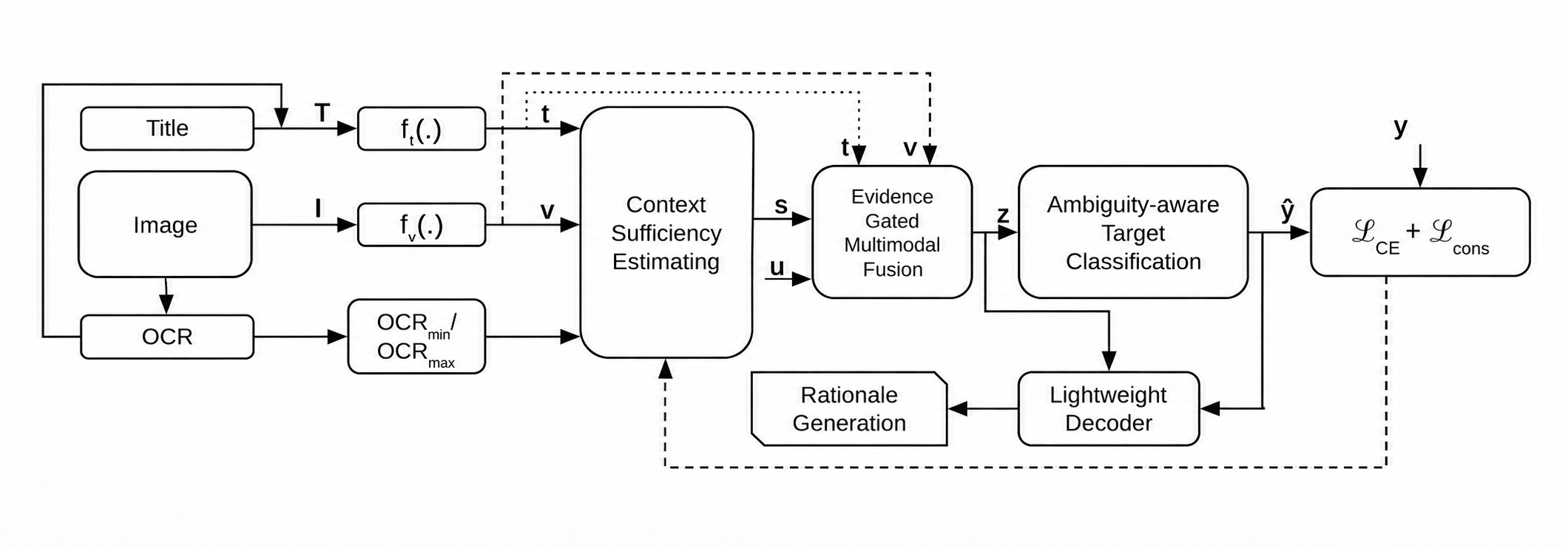}     
\caption{Proposed Framework}
\label{fig:overview}
\end{center} 
\end{figure*}

Biased behavior often emerges when the economic signal in a meme is subtle, implied, or absent altogether. To address this, we present our proposed multimodal framework \textbf{CGER-Net} (\textbf{C}ontext-\textbf{G}rounded \textbf{E}conomic \textbf{R}easoning \textbf{Net}work) to analyze gendered economic dependence in image-text memes when the context ranges from obvious to highly ambiguous. CGER-Net explicitly distinguishes three core components: \textit{evidence extraction}, \textit{context sufficiency} reasoning, and \textit{explanation generation}. The framework enforces epistemic restraint by modeling whether the available multimodal evidence justifies gendered economic attribution and producing justifications that reflect this assessment, as opposed to assuming that all memes permit confident social inference. An overview of the framework is illustrated conceptually in Figure~\ref{fig:overview}.

The proposed architecture consists of five components:  
(i) multimodal economic evidence encoding,  
(ii) context sufficiency estimation,  
(iii) evidence-gated fusion under uncertainty,  
(iv) ambiguity-aware target classification, and  
(v) rationale generation conditioned on evidential sufficiency.

\subsection{Problem Formulation}



Each meme consists of an image $I$ associated with its text $T$, OCR-extracted text, and metadata when available. The task is to assign one label
\[
\hat{y} \in \{\text{Men}, \text{Women}, \text{Neutral}, \text{Ambiguous}\},
\]

\textbf{this}

where \textbf{labels correspond} to the \textit{\textbf{implied subject}} of economic dependence or the absence of sufficient evidence for such attribution.

In contrast to standard bias classification tasks, the Ambiguous label reveals epistemic insufficiency rather than neutrality: the meme lacks sufficient grounded cues to support a gendered economic interpretation. The model not only predicts $\hat{y}$, but it also generates a rationale in natural language, $r$, that explains why the prediction is epistemically justified in light of the multimodal evidence that is available.


\subsection{Multimodal Economic Evidence Encoding}
\label{sec:encoding}

\paragraph{Visual Encoding.}
Given an input image $I$, a pretrained vision encoder $f_v(\cdot)$ produces a fixed-dimensional visual representation:
\[
\mathbf{v} = f_v(I), \quad \mathbf{v} \in \mathbb{R}^{d_v}.
\]
During training, we freeze the encoder to prevent dataset-specific amplified correlations between visual representations and gendered economic roles.

\paragraph{Textual Encoding.}
We encode text $T$ using a pretrained language encoder $f_t(\cdot)$:
\[
\mathbf{t} = f_t(T), \quad \mathbf{t} \in \mathbb{R}^{d_t}.
\]

While gender is not explicitly specified, this representation captures explicit linguistic references to work, finances, dependency, or social roles. We also ensure that no pre-existing cross-modal assumptions are made during the encoding stage by independently operating both visual and textual encoders.





\subsection{Context Sufficiency Estimation}
\label{sec:sufficiency}

A key novelty of CGER-Net is the explicit estimation of whether the available multimodal cues are sufficient to justify gendered economic reasoning. We define a \emph{context sufficiency score}:
\[
s = \sigma\!\left(g([\mathbf{v}; \mathbf{t}; \ell_{\text{ocr}}])\right), \quad s \in [0,1],
\]
where $[\cdot;\cdot]$ denotes concatenation, $\ell_{\text{ocr}}$ is the normalized OCR text length, and $g(\cdot)$ is a lightweight multilayer perceptron. The sigmoid function $\sigma(\cdot)$ maps the output to a probability-like score.

The sufficiency score reflects the degree to which economic dependence cues are explicit and grounded (see, Appendix~\textbf{Limitations}).  Low values of $s$ indicate underspecification or symbolic ambiguity, while high values indicate that the meme contains sufficient evidence for a confident interpretation. This module only determines whether inference is epistemically justified rather than predicting gender.

\subsection{Evidence-Gated Multimodal Fusion}
\label{sec:fusion}

To prevent stereotype-driven inference under insufficient context, CGER-Net employs evidence-gated fusion. Let $\phi(\cdot)$ denote a fusion network operating over concatenated visual and textual representations. The final fused representation is defined as:
\[
\mathbf{z} = s \cdot \phi([\mathbf{v}; \mathbf{t}]) + (1 - s) \cdot \mathbf{u},
\]
where $\mathbf{u}$ is a learned uncertainty embedding, indicates the absence of reliable economic evidence.

It ensures that when context sufficiency is low, both predictions and explanations are conditioned on uncertainty rather than latent correlations. In contrast, the fused representation preserves informative multimodal signals when there is sufficient evidence.



\subsection{Ambiguity-Aware Target Classification}
\label{sec:classification}

The fused representation $\mathbf{z}$ is passed to a classifier producing a probability distribution over the four target categories:
\[
p(\hat{y} \mid I, T) = \text{softmax}(W\mathbf{z} + b).
\]

Training uses a standard cross-entropy loss:
\[
\mathcal{L}_{\text{CE}} = -\sum_{c} y_c \log p(\hat{y}_c),
\]
where $y_c$ denotes the ground-truth label.

To discourage unjustified gender attribution under low sufficiency, we introduce a consistency regularization term:
\[
\mathcal{L}_{\text{cons}} = \mathbb{E}\!\left[(1 - s)\cdot \mathbb{I}(\hat{y} \in \{\text{Men}, \text{Women}\})\right],
\]
which penalizes confident gendered predictions when contextual evidence is weak.

\subsection{Rationale Generation}
\label{sec:rationale}


CGER-Net produces a natural-language rationale $r$ based on both the fused representation $\mathbf{z}$ and the predicted label $\hat{y}$. These rationales are generated by a lightweight decoder that either (i) articulates grounded economic evidence when $s$ is high, or (ii) explicitly states that the available evidence is insufficient when $s$ is low. In this way, the rationales are epistemically aligned with the model’s own sufficiency judgment, rather than serving as post-hoc justifications.


\subsection{Training and Inference}

The final training objective is:
\[
\mathcal{L} = \mathcal{L}_{\text{CE}} + \lambda\,\mathcal{L}_{\text{cons}},
\]
where $\lambda$ controls ambiguity-aware regularization. All pretrained encoders remain frozen, while AdamW is used to train the sufficiency estimator, fusion module, uncertainty embedding, classifier, and rationale decoder. During inference, predictions and explanations are generated using sufficiency-aware gating without ground-truth ambiguity labels, enabling the model to distinguish ambiguous, neutral, and economically biased content.




\section{Experimental Setup}
\label{sec:experimental}

\textbf{CGER-Net} is evaluated on \textbf{EconMeme-GE} to assess gendered economic dependence and uncertainty regarding the gendered meme target, \textit{i.e.}, epistemic ambiguity. Two principles are followed in the design of the evaluation scheme: (i) accurate measurement of biased overcommitment with respect to gender when the available evidence is insufficient (ii) item Ensuring that the observed commitment does not arise from trivial abstention, label collapse, or conservative prediction bias.

\subsection{Baselines}
\label{sec:baselines}
A set of popular and recent Vision-Language (VL) models, which were used in VL and meme understanding research, was selected as a baseline to evaluate CGER-Net. The following models are widely used in vision-language and meme understanding research. \textbf{CLIP ViT-L/14}~\cite{radford2021learning}, \textbf{BLIP-2}~\cite{li2022blip}, the instruction-tuned variant \textbf{InstructBLIP}~\cite{dai2023instructblip}, \textbf{LLaVA-1.6}~\cite{liu2024llava16}, \textbf{Qwen-VL-Chat}~\cite{bai2023qwenvl}, and \textbf{Debiased CLIP (INLP)}~\cite{ravfogel-etal-2020-null} are included in the set of baseline models selected for evaluation. Debiased CLIP (INLP) performs post-hoc removal of gender-correlated subspaces. To ensure fair evaluation, all baseline methods and the proposed CGER-Net are evaluated under the identical dataset splits. Experimental results indicate that, among all compared baselines, only CGER-Net can explicitly reason about contextual sufficiency and appropriately abstain in ambiguous cases, even when prompted to produce both a target prediction and a brief explanation.

\subsection{Evaluation Metrics}
\label{sec:metrics}

The \textbf{CGER-Net} and all baseline models are evaluated on \textbf{Accuracy (Acc)}, \textbf{Macro-F1}, along with metrics to evaluate ambiguity and bias sensitivity. \textbf{Ambiguity Recall (AR)} and the \textbf{Gender Overcommitment Rate (GOR)} measure the correct identification of ambiguous memes and the proportion of ambiguous instances incorrectly assigned to either women or men, respectively. To appraise the quality of explanation, the \textbf{Rationale Alignment Rate (RAR)} measures consistency between the generated rationale and model predictions. RAR evaluates explanation-decision coherence, and GOR serves as the primary metric for the bias-sensitivity metric.

\section{Results}
\label{sec:results}

\subsection{Quantitative Results}

\begin{table*}[!]
\centering
\label{tab:main_results}
\resizebox{0.6\textwidth}{!}
{
\begin{tabular}{lcccccc}
\hline

\textbf{Model} & \textbf{Type} & \textbf{Acc} & \textbf{F1} & \textbf{AR} & \textbf{GOR$\downarrow$} & \textbf{RAR$\uparrow$} \\
\hline
CLIP ViT-L/14 & VLM & 70.4 & 66.9 & 0.44 & 0.37 & -- \\
BLIP-2 & VLM & 72.1 & 68.3 & 0.48 & 0.31 & -- \\
InstructBLIP & VLM & 73.2 & 69.1 & 0.50 & 0.28 & -- \\
LLaVA-1.6 & LMM & 74.5 & 70.6 & 0.52 & 0.34 & 0.46 \\
Qwen-VL-Chat & LMM & 75.1 & 71.3 & 0.51 & 0.32 & 0.49 \\
InternVL-Chat & LMM & \bf 75.6 & \bf 71.8 & 0.53 & 0.30 & 0.51 \\
Debiased CLIP (INLP) & Bias-aware & 71.3 & 67.0 & 0.47 & 0.33 & -- \\
\hline
\textbf{CGER-Net (Ours)} & \textbf{Epistemic MM} & 73.8 & 70.2 & \textbf{0.63} & \textbf{0.19} & \textbf{0.71} \\
\hline
\end{tabular}
}
\caption{The performance comparison of MLM baseline models with \textbf{CGER-Net} shows that under ambiguity, \textbf{GOR} measures (lower is better) unjustified gendered economic attribution. The consistency between generated rationales and predictions is measured by \textbf{RAR} (with higher values indicating better consistency).}
\label{tab_1}
\end{table*}

Pretrained Multimodal Language Models (MLMs) achieve strong performance on vision-language tasks but often hallucinate and exhibit gender overcommitment under ambiguous contexts. Although instruction tuning improves explanation fluency, it does not fully mitigate stereotype-driven overcommitment.

Compared to baseline models, CGER-Net shows stronger epistemic restraint with the highest ambiguity recall and lowest GOR (Table~\ref{tab_1}). Although slightly less accurate than stronger LLMs, it uniquely combines high rationale alignment with low overcommitment, indicating that its performance stems from explicit ambiguity modeling rather than conservative collapse.

\subsection{Performance Across Contextual Explicitness}

\begin{table}[!ht]
\centering
\label{tab:ocr_analysis}
\begin{adjustbox}{max width=0.41\textwidth}
\begin{tabular}{lcccc}
\hline
\textbf{OCR Length} & \textbf{Acc} & \textbf{GOR$\downarrow$} & \textbf{AR$\uparrow$} & \textbf{RAR$\uparrow$} \\
\hline
Low ($<$40 chars) & 61.2 & 0.58 & 0.28 & 0.62 \\
Medium (40--100) & 67.4 & 0.32 & 0.55 & 0.70 \\
High ($>$100) & 72.1 & 0.14 & 0.74 & 0.81 \\
\hline
\end{tabular}
\end{adjustbox}
\caption{The performance of \textbf{CGER-Net} is compared under varying levels of textual explicitness. It demonstrates that prediction and rationale alignment is governed by \textbf{contextual sufficiency}.}
\label{tab_2}
\end{table}

As shown in Table~\ref{tab_2}, as contextual explicitness increases, \textbf{CGER-Net} exhibits decreasing \textbf{GOR} and increasing RAR, reflecting a transition from abstention to confident attribution. In settings with low text availability, where other \textbf{MLMs} tend to hallucinate, the proposed \textbf{CGER-Net} acknowledges evidence insufficiency, controls overcommitment, and still generates coherent rationales.

\section{Ablation Study}
\label{sec:ablation}

\begin{table}[!ht]
\centering

\begin{adjustbox}{max width=0.45\textwidth}
\begin{tabular}{lcccc}
\hline
\textbf{Variant} & \textbf{Acc} & \textbf{AR} & \textbf{GOR$\downarrow$} & \textbf{RAR$\uparrow$} \\
\hline
Full CGER-Net & 73.8 & 0.63 & 0.19 & 0.71 \\
w/o Sufficiency Module & 69.1 & 0.29 & 0.46 & 0.38 \\
w/o Gated Fusion & 68.5 & 0.34 & 0.41 & 0.44 \\
w/o Consistency Loss & 68.9 & 0.37 & 0.38 & 0.49 \\
\hline
\end{tabular}
\end{adjustbox}
\caption{Ablation of CGER-Net, including rationale alignment.}
\label{tab_3}
\end{table}

\paragraph{Explanation.}
Table~\ref{tab_3} shows the ablation study in which the full CGER-Net is compared with the removal of the Sufficiency Module, Gated Fusion, and Consistency Loss. The removal of these components from full CGER-Net leads to a marginal increase in accuracy (up to 5\%), but the important metrics Ambiguity Recall (AR), Gender Overcommitment Rate (GOR), and Rationale Alignment Rate (RAR) show a degradation of around 50\% each. Context sufficiency module is most important for controlling gender overcommitment and rationale alignment. The key takeaway from this study is that ambiguity modelling is indispensable for explanation faithfulness while maintaining high accuracy. 

\subsection{Human Evaluation of Rationales}
\label{sec:human_eval}
CGER-Net produces both predictions and natural-language rationales. While prediction metrics are evaluated in Section~\ref{sec:results}, this subsection assesses the epistemic appropriateness of both outputs through human evaluation. The focus is on whether rationales are grounded in multimodal evidence under contextual ambiguity rather than linguistic fluency. Only CGER-Net is evaluated, as it is specifically designed to generate evidence-backed rationales.

\paragraph{Evaluation Protocol.}
Test instances were randomly sampled from each meme category (Men, Women, Neutral, and Ambiguous). For each instance, evaluators were shown the meme (image–text pair), the model’s predicted label, and the generated rationale, without access to the ground-truth label. Evaluators provided a binary judgment (\textbf{Aligned}/\textbf{Not Aligned}) based on whether the rationale justified the prediction using the meme content; explicit acknowledgment of \textbf{Ambiguity} in cases of insufficient evidence was counted as \textbf{Aligned}.

\paragraph{Evaluation Metric.}
We define the \textbf{Human Rationale Alignment Rate (HRAR)} as:
\[
\text{HRAR} = \frac{\#\text{Aligned rationales}}{\#\text{Evaluated instances}}.
\]

\paragraph{Results.}
\textbf{HRAR }scores of all four categories are reported in Table~\ref{tab_hrar}. \textbf{CGER-Net} shows strong performance on Ambiguous and a high alignment overall. It entails that instead of hallucinating in the evidence insufficiency model acknowledges it and flags \textbf{Ambigous}. Culturally implicit cues are responsible for lower alignment cases. 

\begin{table}[!]
\centering
\label{tab:human_eval}
\begin{adjustbox}{max width=0.36\textwidth}
\begin{tabular}{lc}
\hline
\textbf{Target Category} & \textbf{HRAR} \\
\hline
Men & 0.74 \\
Women & 0.76 \\
Neutral & 0.81 \\
Ambiguous & 0.85 \\
\hline
Overall & 0.79 \\
\hline
\end{tabular}
\end{adjustbox}
\caption{Human evaluation of CGER-Net rationales. \textbf{HRAR} denotes the proportion of rationales judged by human evaluators as aligned with the available multimodal evidence and the predicted label.}
\label{tab_hrar}
\end{table}

\paragraph{Relation to Automatic Metrics.}
HRAR complements the automatic Rationale Alignment Rate (RAR) by providing external human validation. The consistency between HRAR and RAR indicates that CGER-Net’s explanations are not only internally coherent but also epistemically acceptable to human evaluators.

\section{Qualitative Analysis}
\label{sec:qualitative}
We conduct a qualitative analysis to complement quantitative results by examining model predictions and generated rationales. The analysis aims to: (i) verify that bias reduction arises from principled epistemic restraint rather than indiscriminate abstention, (ii) assess whether rationales are supported by multimodal evidence, (iii) identify ambiguous cases where CGER-Net succeeds but MLMs fail, and (iv) analyze failure cases in prediction and rationale generation. 
\subsection{Qualitative Comparison with Strong Multimodal Baselines}

Table~\ref{tab_5} presents an instance-level comparison between a strong multimodal LLM baseline and CGER-Net. In addition to predicted labels, we report rationales generated by CGER-Net to explain its decisions. Correct grounded predictions are highlighted in \textcolor{green!60!black}{green}, incorrect or biased predictions in \textcolor{red!70!black}{red}, and epistemically appropriate abstentions in \textcolor{blue!70!black}{blue}.

\begin{table*}[!]
\centering
\label{tab:qualitative_extended}
\renewcommand{\arraystretch}{1.32}
\begin{adjustbox}{max width=\textwidth}
\begin{tabular}{p{5.5cm}|p{1.5cm}|p{1.5cm}|p{1.5cm}|p{6.5cm}}
\hline
\textbf{Meme Description (Image + Text)} &
\textbf{GT} &
\textbf{LLaVA-1.6} &
\textbf{CGER-Net} &
\textbf{CGER-Net Rationale (Summary)} \\
\hline

\includegraphics[width=1.4in,height=0.6in]{couple_at_table.jpg}

\vspace{1pt} text: ``So who’s paying this time?''  
(No explicit income or role specified)
&
Ambiguous
&
\textcolor{red!70!black}{Women}
&
\textcolor{blue!70!black}{Ambiguous}
&
The meme implies financial responsibility but does not specify who earns or pays; gender attribution is unsupported by the content. \\
\hline

\includegraphics[width=1.4in,height=0.6in]{women_bags.jpeg}

\vspace{1pt} text: ``My paycheck, my choices''
&
Women
&
\textcolor{blue!70!black}{Ambiguous}
&
\textcolor{green!60!black}{Women}
&
The text explicitly links financial agency to the speaker, who is visually and linguistically identified as a woman. \\
\hline

\includegraphics[width=1.4in,height=0.6in]{men_cartoon.jpeg}

\vspace{1pt} caption: ``Unemployed but still confident''
&
Men
&
\textcolor{green!60!black}{Men}
&
\textcolor{green!60!black}{Men}
&
Economic dependence is explicitly stated through unemployment and the subject is clearly male. \\
\hline

\includegraphics[width=1.4in,height=0.6in]{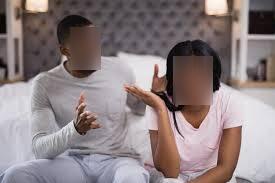}

\vspace{1pt} text: ``Money problems again''  
(No source of income mentioned)
&
Ambiguous
&
\textcolor{red!70!black}{Men}
&
\textcolor{blue!70!black}{Ambiguous}
&
Financial stress is mentioned, but no evidence assigns responsibility or dependence to a specific gender. \\
\hline

\includegraphics[width=1.4in,height=0.6in]{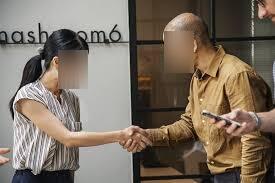}

\vspace{1pt} text: ``Another promotion, another bonus''
&
Women
&
\textcolor{green!60!black}{Women}
&
\textcolor{green!60!black}{Women}
&
Explicit reference to promotion and bonus grounds financial agency in a female subject. \\
\hline

\includegraphics[width=1.4in,height=0.6in]{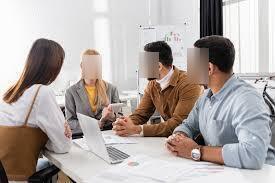}

\vspace{1pt} text: ``Teamwork makes it work''
&
Neutral
&
\textcolor{red!70!black}{Men}
&
\textcolor{green!60!black}{Neutral}
&
Both individuals are depicted as working, with no indication of dependence or financial hierarchy. \\
\hline

\includegraphics[width=1.4in,height=0.8in]{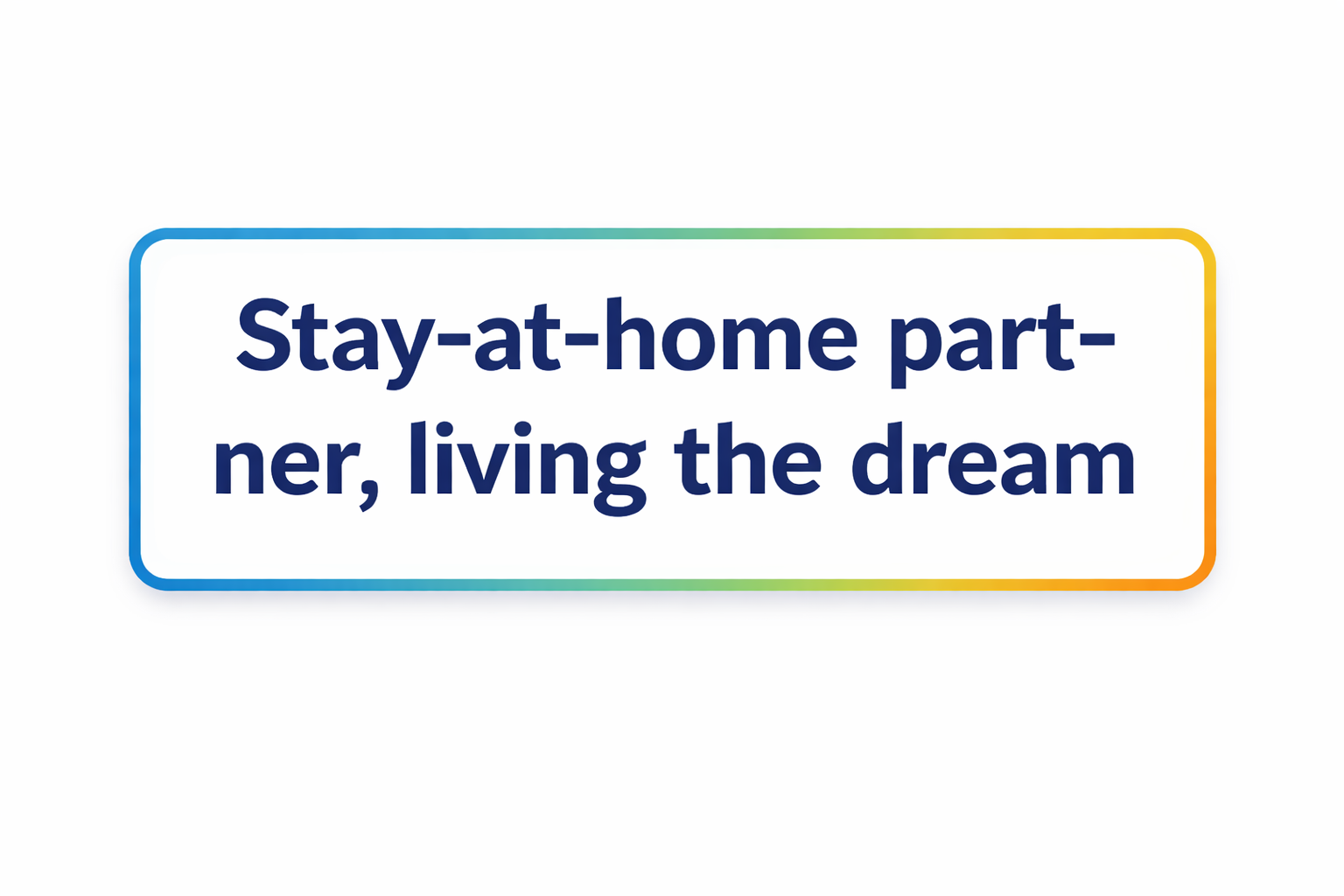}

\vspace{1pt} text: ``Stay-home''  
(No gender specified)
&
Ambiguous
&
\textcolor{red!70!black}{Women}
&
\textcolor{blue!70!black}{Ambiguous}
&
Economic dependence is implied, but the gender of the dependent partner is unspecified. \\
\hline

\includegraphics[width=1.4in,height=0.6in]{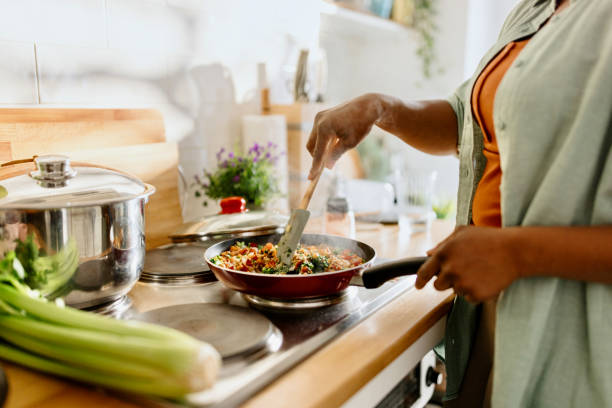}

\vspace{1pt} text: ``Supporting the household in my own way''  
(Economic role indirect)
&
Ambiguous
&
\textcolor{red!70!black}{Women}
&
\textcolor{blue!70!black}{Ambiguous}
&
Domestic contribution is mentioned, but financial dependence or agency is not explicitly stated. \\
\hline

\end{tabular}
\end{adjustbox}
\caption{Qualitative analysis with rationale inspection for gendered economic dependence reasoning. The table contrasts predictions from \textbf{LLaVA-1.6} (a strong multimodal large language model baseline) with CGER-Net outputs and their corresponding rationales. Color coding highlights epistemically appropriate abstention (\textcolor{blue!70!black}{blue}), correct grounded attribution (\textcolor{green!60!black}{green}), and unjustified or incorrect inference (\textcolor{red!70!black}{red}). Rationales are shown in abbreviated form for clarity.}
\label{tab_5}
\end{table*}

\subsection{Why CGER-Net Succeeds Under Ambiguity}

Across ambiguous cases, strong multimodal LLM baselines frequently resolve underspecification by invoking latent gender stereotypes, despite limited or absent evidence. These decisions are often confident but unsupported by explicit economic cues. CGER-Net avoids this failure mode due to two interacting mechanisms: \textbf{Context Sufficiency Estimation}: Weak textual grounding and limited economic cues result in low sufficiency scores, discouraging premature commitment. \textbf{Evidence-Gated Fusion}: Under low sufficiency, the model relies on uncertainty-aware representations, which propagate into both the prediction and the generated rationale.


As reflected in Table~\ref{tab_5}, CGER-Net’s rationales explicitly reference evidential absence rather than hallucinating gendered economic roles, aligning prediction and explanation.

\subsection{When CGER-Net Commits Correctly}

In unambiguous cases, CGER-Net generates confident predictions accompanied by grounded rationales that cite explicit textual or visual evidence (e.g., references to salary, promotion, or unemployment). This demonstrates that the model does not suppress gendered economic reasoning wholesale; instead, it commits selectively when sufficient evidence exists.

\subsection{Failure Analysis: Rationale-Aware Errors}

Despite overall improvements, CGER-Net exhibits conservative failure modes:

\paragraph{Implicit Cultural Semantics.}
Memes where economic dependence is culturally implied (e.g., domestic labor framed as contribution) may receive \textit{Ambiguous} predictions even when annotators assign gendered labels. The corresponding rationales correctly cite lack of explicit financial evidence, indicating epistemic caution rather than misinterpretation.

\paragraph{Visually Dominant Cues with Minimal Text.}
In rare cases, strong visual stereotypes combined with extremely sparse text can influence sufficiency estimation. While CGER-Net reduces such effects, rationales reveal that visual symbolism alone is insufficient to fully ground economic attribution.

\section{Conclusion}

We presented \textbf{CGER-Net}, a context-grounded multimodal framework for analyzing gendered economic dependence in memes. By explicitly modeling contextual sufficiency, enforcing evidence-gated inference, and generating epistemically aligned rationales, CGER-Net substantially reduces unjustified gender attribution under ambiguous economic contexts. Extensive quantitative, qualitative, ablation, and human-centered evaluations on the \textbf{EconMeme-GE} dataset demonstrate that the proposed approach mitigates bias without suppressing valid economic reasoning when evidence is explicit. Our findings underscore the importance of epistemic restraint and explanation alignment for the responsible deployment of multimodal AI systems in socially sensitive domains.

\section*{Limitations}
The current framework relies on OCR length and surface-level textual cues as proxies for contextual richness, which may not capture subtle economic symbolism. Additionally, cultural variability in interpreting economic roles may affect both annotation and model behavior. Future work will explore culturally adaptive sufficiency estimation and richer rationale supervision.

\appendix
\section*{Dataset Details}
\label{appendix_dataset}
\subsection*{Annotator Demographics and Inter-Annotator Agreement}
\label{ap_and_agreement}
All three annotators (2 NLP experts and 1 senior annotator who works on gender and economic studies) are from South Asian backgrounds. Annotation guidelines (Section~\ref{anno_guidelines}) explicitly instruct avoiding stereotype-driven reasoning. Despite shared cultural context but different areas of research, Cohen's $\kappa$\,=\,0.74 indicates substantial agreement. The primary disagreement axis was culturally implicit economic cues (e.g., domestic labor framed as contribution), as discussed in Section~\ref{annoation_challenge}.

EconMeme-GE contains a total of \textbf{2,048} image--text memes curated from publicly available social media sources. The dataset is intentionally balanced across explicit and underspecified economic contexts to support robust evaluation under ambiguity. Approximately \textbf{41\%} of the instances are labeled as \textit{Ambiguous}, reflecting the prevalence of implicit or insufficiently grounded economic cues in real-world social content. The remaining instances are distributed across \textit{Men} (\textbf{22\%}), \textit{Women} (\textbf{24\%}), and \textit{Neutral} (\textbf{13\%}) categories. On average, memes contain \textbf{47} OCR-extracted characters, with \textbf{38\%} of instances containing fewer than 40 characters, highlighting the sparsity of textual evidence that motivates epistemic abstention. All images are naturalistic social media visuals, and no external demographic or identity metadata is provided. The dataset is split into \textbf{70\%} training, \textbf{10\%} validation, and \textbf{20\%} test sets, stratified by label to preserve class distributions across splits.

\subsection*{Dataset Availability}
\label{dataset_avilability}
The dataset, along with the search queries used for web-based data collection and dataset construction, will be publicly released at \url{https://github.com/kushalkanwarNS/Gendered_Economic_Memes}.

\bibliographystyle{named}
\bibliography{ijcai26}

\end{document}